\documentclass[authoryear,review,12pt,pdf]{elsarticle}
\usepackage{cite}
\usepackage{adjustbox}
\usepackage{amssymb}
\usepackage{amsmath}
\usepackage{array}
\usepackage{blindtext}
\usepackage{booktabs} 

\usepackage[labelfont=bf]{caption}
\DeclareCaptionLabelFormat{mycaption}{#1 #2.}
\usepackage{color}
\usepackage[dvipsnames]{xcolor}
\usepackage{csquotes}
\usepackage{float}
\usepackage{geometry} 
\usepackage{graphicx}
\usepackage[colorlinks=true,linkcolor=red, citecolor=blue, urlcolor=blue,pdfauthor=author]{hyperref}
\usepackage{lipsum}
\usepackage{lscape}
\usepackage{lineno}
\usepackage{mathtools}
\usepackage{makecell}
\usepackage{multirow}
\usepackage{pifont}
\usepackage{pdflscape}
\usepackage{rotating}
\usepackage{soul}
\setstcolor{red}
\usepackage{tabularx,booktabs}
\usepackage{threeparttable}
\usepackage{textcomp}
\usepackage{tikz} \usetikzlibrary{shapes,arrows}
\usepackage{url}
\usepackage{varioref}
\usepackage{wrapfig}
\usepackage[dvipsnames]{xcolor}
\definecolor{dblue}{rgb}{0.16,0.49,0.66}
\definecolor{dgreen}{rgb}{0.0,0.69,0.313}
\usepackage{siunitx}
\usepackage{color}
\newcolumntype{P}[1]{>{\centering\arraybackslash}p{#1}}
\newcolumntype{M}[1]{>{\centering\arraybackslash}m{#1}}

\DeclareMathAlphabet {\mathbfit}{OML}{cmm}{b}{it}

\begin{document}

\begin{frontmatter}
\title{A Deep Learning Architecture for Passive Microwave Precipitation Retrievals using CloudSat and GPM Data}
\author[add1]{Reyhaneh Rahimi}
\author[add1]{Sajad Vahedizade} 
\author[add1]{Ardeshir Ebtehaj \corref{cor1}}\ead{ebtehaj@umn.edu}

\cortext[cor1]{Corresponding Author}

\address[add1]{Saint Anthony Falls Laboratory and Department of Civil Environmental and Geo- Engineering, University of Minnesota, Minneapolis, MN, 55455 USA.}


\begin{abstract}
This paper presents an algorithm that relies on a series of dense and deep neural networks for passive microwave retrieval of precipitation. The neural networks learn from coincidences of brightness temperatures from the Global Precipitation Measurement (GPM) Microwave Imager (GMI) with the active precipitating retrievals from the Dual-frequency Precipitation Radar (DPR) onboard GPM as well as those from the {\it CloudSat} Profiling Radar (CPR). The algorithm first detects the precipitation occurrence and phase and then estimates its rate, while conditioning the results to some key ancillary information including parameters related to cloud microphysical properties. The results indicate that we can reconstruct the DPR rainfall and CPR snowfall with the detection probability of more than 0.95 while the probability of false alarm remains below 0.08 and 0.03, respectively. Conditioned to the occurrence of precipitation, the unbiased root mean squared error in estimation of rainfall (snowfall) rate using DPR (CPR) data is less than 0.8 (0.1) \si{mm.hr^{-1}} over oceans and land. Beyond methodological developments, comparing the results with ERA5 reanalysis and official GPM products demonstrates that the uncertainty in global satellite snowfall retrievals continues to be large while there is a good agreement among rainfall products. Moreover, the results indicate that CPR active snowfall data can improve passive microwave estimates of global snowfall while the current CPR rainfall retrievals should only be used for detection and not estimation of rates.  
\end{abstract}

\end{frontmatter}

\section{Introduction}
Precipitation processes play a vital role in space-time dynamics of global water and energy cycle \citep{Kidd2011, Levizzani2019}. Spatiotemporal variability of precipitation and its phase change affect availability of water resources to ecosystem and society -- impacting decision making across a wide range of socioeconomic sectors, especially under a changing climate \citep{Tamang2020, Qin2021}. Ground-based rain gauges and precipitation radars are not available over oceans and hardly provide a global picture of overland variability of precipitation. These networks are sparse, especially over remote areas with complex terrains, and their measurements are subject to a large-degree of uncertainty due to variety of reasons such as wind-induced under-catch losses of rain gauges \citep{Fuchs2001, Duchon2010} or topographic blockage of ground-based radars. Remote sensing from space has promised to cope with these shortcomings and provide a quasi-global picture of precipitation variability \citep{Rasmussen2012, Guilloteau2021}; however, there are still technical and knowledge gaps that need to be addressed -- especially with respect to monitoring of global snowfall.  

Observations provided by passive and active sensors onboard the satellites have been instrumental for almost four decades to expand our understanding of the role of precipitation and its variability in the global hydrologic cycle \citep{Olson1996, Kummerow1998, Grecu2006, Sorooshian2011, Kirschbaum2017, Skofronick-Jackson2017}. The passive microwave (PMW) brightness temperatures (TB) with wavelengths ranging from 10 to 200 GHz are sensitive to the columnar integrated precipitable water content \citep{Petty1994, Skofronick-Jackson2004, Kummerow2001}. Over radiometrically cold oceans, the observed TBs increase in response to raindrop emission at frequencies below the 60 GHz oxygen absorption line. Consequently, low-frequency channels $\leq$37 GHz have been widely used for detection of rainfall over oceans \citep{Bauer2001, Kummerow2001, Petty_2013_a, EbtBF15}. Over higher frequencies, where wavelengths approach to the size of the ice particles, the observed TBs decrease as the upwelling surface emission is scattered by the atmospheric constituents -- especially ice particles. Therefore, frequencies above 80 GHz have been employed to detect ice clouds and snowfall \citep{Bennartz2001,Bennartz2003,Petty2010, Skofronick_Jackson2011}.

A more direct measurement of precipitation microphysics can be obtained from active precipitation radars, which rely upon measurements of the backscattered power of the transmitted electromagnetic waves due to volume scattering of hydrometeors \citep{Iguchi2012, Toyoshima2015, Grecu2016, Heymsfield2018}. Among existing satellites, the Global Precipitation Measurement (GPM, \citep{Hou2014}) core satellite launched in 2014 as well as the CloudSat satellite (2006-present, \citep{Stephens2002}) have provided frequent passive and active observations of near-global precipitation.

The GPM core satellite carries a microwave imager (GMI, 10--187 GHz) and a Ku-Ka (13-35 GHz) band Dual-Frequency Precipitation Radar (DPR) \citep{Skofronick-Jackson2017} with a coverage of 65$^\circ$ S-N. The GMI scans an outer swath width of 931~km with a mean footprint resolution ranging from 25~km at 10.65 GHz to 6~km at 183.3$\pm$7 GHz channels \citep{Draper2015}. The effective field of view of the radiometer varies from 19.4$\times$32.1~km to 3.8$\times$5.8~km (along-scan$\times$cross-scan) at 10.65 GHz and 183.31$\pm$7 GHz, respectively. The DPR provides temporally less frequent reflectivity values with an approximate footprint of 5~km resolution over a swath width of 245~km at both Ka and Ku bands. 

On the other hand, the W-band (94 GHz) Cloud Profiling Radar (CPR, 2006-present) onboard the CloudSat \citep{Stephens2002, LEcuyer2010} enables probing the cloud and light precipitation profiles with reflectivity values as low as -30~dBZ \citep{Liu2008a} with an effective vertical resolution of 240~m over a 1.8$\times$1.4~km (along-track$\times$cross-track) spatial resolution within 82$^\circ$ S-N. Therefore, despite extremely low temporal resolution of CPR, it can be considered as one of the most accurate spaceborne instruments for sensing light precipitation events \citep{Behrangi2016, Kulie2018, Bennartz2019}. 


Coincident observations of both active radars and passive microwave radiometers have been used to develop PMW precipitation retrieval algorithms \citep{Kummerow1996, Skofronick-Jackson2003c, Grecu2004, Kummerow2011, Ebtehaj_2016_eval,Turk2018, Munchak2020}. Coincidences of GMI and DPR have been commonly used in the context of Bayesian inference models for PWM precipitation retrievals \citep{Kummerow2001, Petty2013a, Kummerow2015, Grecu2016, Ebtehaj2017}. The Bayesian algorithms often match the observed TBs with an \textit{a priori} database that links a statistically representative number of TBs to their radar precipitation profiles. This matching enables to decide about the occurrence and phase of precipitation in a probabilistic sense. However, recent research suggested that DPR is not sufficiently sensitive in capturing backscattering signatures of light and shallow precipitations as it can only record those events with reflectivity above 12-18~dBZ \citep{Liu2008a, Hamada2016}, which roughly correspond to precipitation rate of 0.2-0.5 \si{mm.hr^{-1}} \citep{Skofronick-Jackson2013, You2017, casella2017evaluation}.

The W-band CPR can properly capture cloud ice particles, droplets, and snowflakes with reflectively values greater than -30~dBZ. Recently, a body of research has been devoted to developing precipitation and snowfall retrieval algorithms through exploiting GMI and CPR coincidences \citep{Rysman2018, Rysman2019, Vahedizade2021, Turk2021}. However, it is important to highlight that as precipitation intensity increases, the CPR reflectivity tends to saturate near 20~dBZ as the size of the hydrometeors becomes significantly larger than the wavelength at W-band \citep{Matrosov2009}. This saturation often increases the uncertainty of the retrievals as the rain rate becomes higher than 10~\si{mm.hr^{-1}} \citep{LEcuyer2002, Tang2017}.

In parallel to the Bayesian retrieval algorithms, artificial neural networks (ANN) have been deployed for PMW precipitation retrievals largely from infrared satellite observations \citep{Sorooshian2000, Hong2004, Tapiador2004}, and to a lesser extent, from microwave bands \citep{Liou1999, Aires2001, Sano2016, Sano2018}. These ANNs are learning architectures that propagate the input data across a sequence of weighted averaging followed by non-linear operations in a series of layers. To properly adjust the network weights based on the training dataset, a learning algorithm computes gradient vectors that sequentially capture the steepest descent direction towards a (local) minimum of a properly designed cost function. Predictive capabilities of classic ANNs were limited due to two main reasons. First, those networks could not be scaled up to solve large-scale problems due to their fully connected architecture. Second, classic optimization algorithms could not update the network weights for high-dimensional data and where the number of layers and computational nodes could be staggeringly large \citep{Glorot2010}.

In the last decade, there have been significant advances in learning paradigms including stochastic \citep{Bottou1991} and batch gradient \citep{Dekel2012} approaches that efficiently approximate the gradient vector and reduce its dependence on the size of the network and the initial weights. Modern non-degenerative activation functions such as the rectified linear units (ReLU) \citep{Nair2010} were introduced to avoid vanishing gradient problem and to promote sparsity in the weights. Moreover, to overcome over-fitting, new effective and computationally amenable regularization methods, such the Dropout technique \citep{Srivastava2014}, were proposed. These advances, as well as high-capacity graphical processing units (GPU) have made it possible to increase the number of hidden layers significantly -- coining the term ``deep learning'' -- and expand the predictive accuracy of the neural network beyond what was possible in shallow ANNs. Different deep neural networks (DNN) have been developed in recent years \citep{LeCun2015, Wang2017, Goodfellow2017}, solving numerous problems in signal and image processing such as object detection \citep{Deng2018, Zhao2019}, image classification \citep{Li2018, Li2019}, and speech recognition \citep{Deng2013, Nassif2019}) with an unprecedented accuracy. 

Recently, there has been a renewed interest to advance PMW retrievals of precipitation and other atmospheric constituents through DNN models. In one of the early attempts \citep{Pfreundschuh2018}, a Quantile Regression Neural Networks (QRNN) was developed to estimate cloud physical properties using observations by the Moderate Resolution Imaging Spectroradiometer (MODIS, \citep{Platnick2003}). In \citep{Tang2018}, a shallow multilayer perceptron (MLP) network with two hidden layers was proposed that used reanalysis data as well as coincidences of GMI TBs with both DPR and CPR retrievals of precipitation to obtain passive estimates of high latitude rain and snow. The proposed architecture used the tangent sigmoid as an activation function and the Levenberg--Marquardt algorithm \citep{More1978} to update the network weights. A deep MLP network was employed in \citep{Chen2020} to retrieve the rainfall information using both infrared and PMW data and its performance was validated over the Dallas--Fort Worth metroplex, Texas, United States. More recently, to extract spatial features of precipitation, a U-Net convolutional neural network was developed \citep{Gorooh2022} -- allowing high resolution retrievals through ingesting data from infrared and PMW satellites over different land-cover types.

Building upon previous research, this paper aims to answer the following key questions: Can DNN architectures lead to PMW retrievals with an accuracy beyond the existing Bayesian algorithms that rely on $k$-nearest matching \citep{EbtBF15,Takbiri2019, Vahedizade2021}? How can physically relevant variables such as cloud water content be incorporated in the retrievals and what would be their effects on the uncertainties? Where and to what extent can CPR data lead to improved retrieval of precipitation at different phases beyond those learned from DPR data? Can deep neural networks disentangle the precipitation and snow-cover microwave signatures without any given information about the presence of snow on the surface?   

To answer the above questions, the paper has the following contributions. (i) This study uses a pixel-level dense DNN architecture based on two databases of coincidences of GMI--DPR as well as GMI--CPR. (ii) The proposed architecture has sequential detection and estimation modules to first identify the precipitation occurrence and its phase (i.e., rain vs snow) and then estimate its rate. (iii) Physical variables including clouds liquid and ice water path, total precipitable water, 2-m air temperature, and convective potential energy (CAPE) are used along with the TB values for detection of precipitation and its phase through a multinomial classification scheme. (iv) The power of the DNN and {\it k}-nearest neighbor matching is combined to promote localization in estimation of precipitation rate. (v) The DNNs employ ReLU activation functions \citep{Nair2010}, batch normalization \citep{Ioffe2015}, Dropout regularization \citep{Srivastava2014}, and the Root Mean Squared Propagation (RMSProp) \citep{Hinton2012} algorithm to update the network weights using mean squares (absolute) error loss functions for retrieval of rainfall (snowfall). 

The paper is organized as follows. Section~\ref{sec:2} describes the data including the GPM and CloudSat products as well as the ERA5 reanalysis data from the European Centre for Medium-Range Weather Forecasts (ECMWF) \citep{ERA5}. Section~\ref{sec:3} explains the methodology. The results, findings, and discussions are presented in Section~\ref{sec:4}. In this section, retrievals and the performance of the proposed DNN are presented and compared with a Bayesian approach \citep{EbtBF15} for training based on both DPR and CPR data. Section~\ref{sec:5} provides a summary, concluding remarks, and points out to future research directions.

\begin{figure*}[t]
    \centering
    \includegraphics[width=1\textwidth]{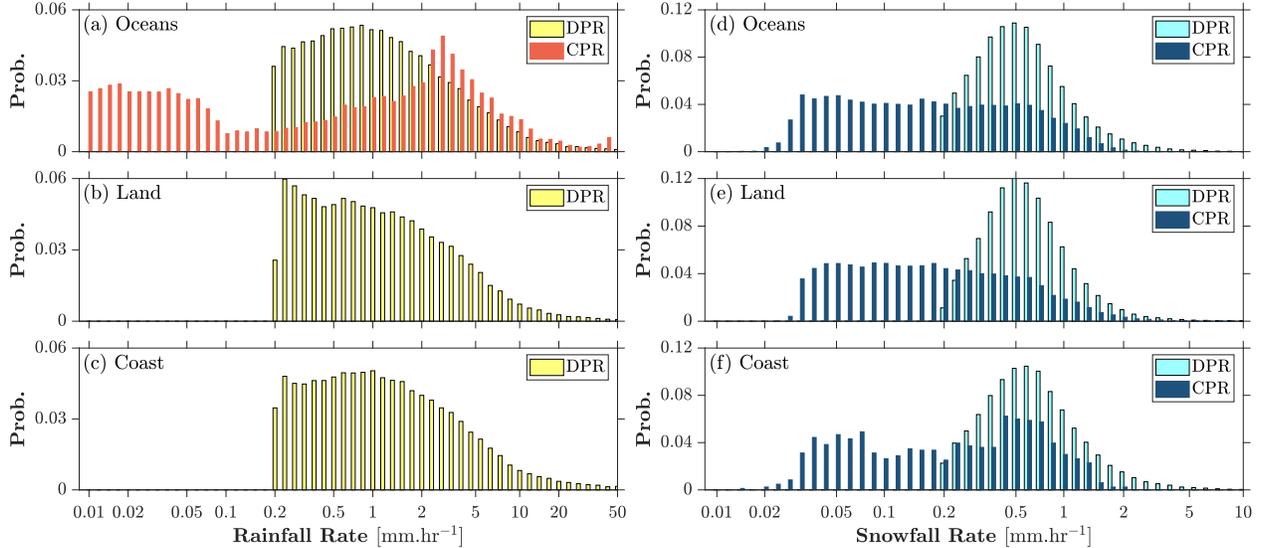}
    \caption{Probability histograms of the GPM DPR and {\it CloudSat} CPR precipitation over three different surface types defined in the GPROF Algorithm Theoretical Basis Document (ATBD, \citep{GPROF}). }
    \label{fig:01}
\end{figure*}

\section{Data} \label{sec:2}

In this study, we use near-coincident observations of GMI and CPR from March 2014 to August 2016 \citep{Turk2021}, coincidences of GMI and DPR data in 2015, as well as the ERA5 reanalysis data from the European Centre for Medium-Range Weather Forecasts \citep[ECMWF,][]{ERA5}. Note that the samples size of CPR coincidences is much smaller than the DPR and thus we need more years to collect adequate CPR data for robust training of the algorithm. The Multi-Radar/Multi-Sensor System (MRMS) radar precipitation \citep{Kirstetter2012, Zhang2016} as well as precipitation products from the Goddard Profiling Algorithm \citep[][V07]{Kummerow2015} are used for cross comparison and retrieval assessment.

The GMI-CPR coincidences rely on the level-II CloudSat products (R05) including the 2C-PRECIP-COLUMN \citep{Haynes2009} and the 2C-SNOW-PROFILE \citep{Wood2013, Wood2014} that respectively contain near-surface rain and snowfall rates. To avoid ground-clutter contamination over complex elevated terrains, the near-surface snowfall rate is reported at 3rd (5th) radar bin above the oceans (land) at 720\,\si{m} (1200\,\si{m}) above the surface\citep{Kulie2010}. In these coincidences, rainfall retrievals over land and coastal areas are not available as they were not reported in the original CloudSat products.  

The 2A-DPR product \citep{Iguchi2018}, retrieved from Ku-band reflectivity, includes the near-surface precipitation phase and its intensity. Furthermore, ancillary information of 2-m air temperature, cloud liquid and ice water paths, and convective available potential energy (CAPE) are added to all coincidences using the ERA5 reanalysis products \citep{ERA5} -- available at spatial resolution of 31~km. It should be noted that the DPR algorithm uses precipitation phase at the lowest radar range gate uncontaminated by surface clutter, which may be 0.5--2.0 km above the surface (even over oceans). The profile of phase (ice, melting, or liquid) is determined via a bright band detection algorithm using the temperature profile provided by ancillary data Japan Meteorological Agency's Global Analysis \citep[GANAL,][]{JMA}. Therefore, phase changes between the actual surface and DPR “near surface” are definitely possible \citep{skofronick2019satellite}. Throughout, the GMI--DPR and GMI--CPR databases are divided based on different surface types including ocean, land, and coastal zones using the information provided by the Goddard Profiling Algorithm \citep[GPROF,][]{GPROF}. 

In summary, the {\it a priori} database contains 40E+6 (5.5E+6) number of GMI-DPR (GMI-CPR) coincidences over oceans. The number of coincidences are 6.5E+6 (2E+6) over land and 2E+6 (6E+5) over coastal zones. These databases will be used to train, validate, and test the proposed retrieval algorithm. The GMI observations are from the calibrated TBs (1C.GPM.GMI, V05) \citep{Berg2016}. The coincidences with DPR are reported at 5~km spacing where both S1 (10--89 GHz) and S2 (166--183 GHz) GMI channels have overlapping observations. 

Probability histograms of the collected CPR and DPR precipitation data at two different precipitation phases are shown over the three surface types in Fig.~\ref{fig:01}. The CPR rainfall data over ocean represent a bimodal distribution. The dominant mode of the distribution is around 3~\si{mm.hr^{-1}}, which is $\sim$2~\si{mm.hr^{-1}} greater than the mode of the DPR data. The distribution of CPR snowfall is much wider than DPR and has a significantly smaller mean values $\sim$0.2~\si{mm.hr^{-1}} across all surface types than the DPR counterpart with a mean around 0.5~\si{mm.hr^{-1}}.

\section{Methodology} \label{sec:3}

Focusing on the PMW precipitation retrievals, we briefly review the basics of the classic Bayesian retrieval algorithms. Then, we elaborate upon the design of the proposed DNN model and its connection with a particular Bayesian model presented in \citep{EbtBF15}.    

\subsection{Bayesian Retrieval Algorithms}

Atmospheric radiative transfer (RT) models \citep{Liu1993,Evans1995,Liu1997} require a large number of parameters and input variables of atmospheric column as well as spectral surface emissivity values to simulate the TBs at top of the atmosphere. Inversion of these forward models is often severely ill-posed as precipitation rate becomes non-uniquely related to the observed TBs at a few frequency channels. To cope with this non-uniqueness, a Bayesian class of inversion models has emerged \citep{Kummerow2001,Kummerow2011,Petty2013b,You2015,Ebtehaj2015,Ebtehaj2016} that learns from an \textit{a priori} database of precipitation profiles and their corresponding simulated/observed TBs. 

Conceptually, given a pixel-level vector of TBs, a Bayesian algorithm attempts to isolate a subset of physically similar vectors of TBs in the database and retrieve statistics of their associated precipitation profiles. Physically similar TBs can be isolated by  subsetting the database based on the known underlying physical conditions \citep{Kummerow2015,You2015}, low-dimensional matching of pseudo channels \citep{Petty2013a,Turk2018}, and/or through a multi-frequency $k$-nearest neighbor ($k$NN) matching \citep{EbtBF15,Ebtehaj_2016_eval, Takbiri2019,Guilloteau2020,Ebtehaj2020,Vahedizade2021}. Here we provide a brief description of the $k$NN approach that we will combine its localized linear embedding properties with the proposed deep learning architecture. 

In particular, let us assume that a collection of $M$ vectors of TBs, namely $\mathbf{T}_{b} \in {\mathbb{R}^{n}}$, at $n$ frequency channels, and their corresponding precipitation properties $\mathbf{p} \in {\mathbb{R}^{3}}$, including the near-surface precipitation occurrence, phase, and rate are collected in a database $\mathcal{B}_{\ell} = \{(\mathbf{T}_{b_{i}}, \mathbf{p}_{i})\}^{M} _{i=1}$, over the $\ell$\textsuperscript{th} surface type that is either ocean, land, or coast. Given an observed pixel-level vector of $\mathbf{T}^{\text{obs}}_b$ over the surface type $\ell$, its $k$-nearest neighbors $k\ll M$ will be identified in $\mathcal{B}_{\ell}$ using the Euclidean or the Mahalanobis distance \citep{Ebtehaj2020}.

The isolated subset $\{\mathbf{T}_{b_i}^\mathfrak{k}\}_{i=1}^k$ of $k$-nearest neighbors and their corresponding precipitation properties $\{\mathbf{p}_{i}^\mathfrak{k}\}_{i=1}^k$ can be used to {\it detect} the occurrence and phase of precipitation associated with $\mathbf{T}^{\text{obs}}_b$ through a nested majority vote rule. Specifically, if majority of $\{\mathbf{p}_{i}^\mathfrak{k}\}_{i=1}^k$ are precipitating, $\mathbf{T}_b^{\text{obs}}$ will be labeled as precipitating. The phase is then decided through a similar majority vote rule among the neighboring precipitating profiles. 

Given the precipitation occurrence and phase of $\mathbf{T}^{\text{obs}}_b$, the inverse model linearly combines the precipitating neighbors with the same precipitation phase (i.e., snow vs. rain) to {\it estimate} the near-surface precipitation rate $\hat{r} = \sum_{i=1}^{k^{\prime}} \mathcal{\omega}_{i} \times r_{i}^{\mathfrak{k}}$, where $r_i^{\mathfrak{k}}$ denotes the $i^{th}$ neighboring near-surface rate, $w_i$ is a weight, and $k^{\prime}\leq k$ is the number of precipitating profiles with the same phase. The weights need to sum to one and can be obtained either through a look-up table \citep{Petty2013b}, an inverse distance weighting interpolation \citep{Kummerow2001} or a constrained regularized least-squares \citep{Ebtehaj2015}. 

\subsection{DIEGO: Deep-learning precIpitation rEtrieval alGOrithm}

\subsubsection{Basics of the Design and Architecture}

As briefly explained, a DNN is a universal approximator that establishes a functional relationship between any learnable set of input-output data points \citep{Atkinson1997} through interconnected neural nodes in multiple layers as discussed in numerous seminal works \citep{Rosenblatt1958, Bishop2006, Bengio2009, LeCun2015, Goodfellow2017}. Here we briefly explain this structure tailored to the PMW precipitation retrieval problem for the sake of completeness. Note that we call the used neural networks ``deep" not solely because they have 4 to 6 layers and more than 200 neurons. The reason is that we use modern training and regularization techniques that enable to propagate the information content of data into a large number of layers and train the network with an accuracy beyond traditional methodologies.

Let us assume that a neural network consists of $l=1,\ldots,L$ layers, including the input and output layers, where $l-1$ and $l$\textsuperscript{th} layers indicate two consecutive hidden layers with $n$ and $m$ neurons, respectively. Consider that $\mathbf{a}^{[l-1]}\in\mathbb{R}^m$ is the output of all neurons in layer $l-1$, $\mathbf{W}^{[l-1]}\in\mathbb{R}^{n\times m}$ denotes the weights connecting the layer $l-1$ to the layer $l$, and $\mathbf{b}^{[l]}\in\mathbb{R}^m$ contains the biases in layer $l$. A fully connected architecture uses repeated application of a nonlinear activation function $\sigma(\mathbf{W}^{[l-1]}\mathbf{a}^{[l-1]}+\mathbf{b}^{[l]}): \mathbb{R}^m \xrightarrow{} \mathbb{R}^m$ for all neurons over each layer and pass their outputs to all neurons in the next layer.    

Let us assume that we have $M$ input-output training data points $\{\mathbf{T}_{b_i}^\mathfrak{c},\, \mathbf{t}_i,\,r_i\}_{i=1}^M$. Here, each $\mathbf{T}_{b_i}^\mathfrak{c}\in\mathbb{R}^{n+n_s}$ denotes a \textit{concatenated} set of inputs that contains TBs at $n$ frequency channels as well as $n_s$ physically relevant state variables (e.g., 2-m air temperature); $\mathbf{t}_i$ are the labels denoting the state of the atmosphere near the surface (i.e., non-precipitating, raining, snowing); and $r_i$ is the surface precipitation rate. In this context, for example the outputs from the last layer of a DNN with four layers has the following functional form:
\[
    \mathcal{F}(\mathbf{T}_{b_i}^\mathfrak{c})= \sigma(\mathbf{W}^{[4]}\sigma(\mathbf{W}^{[3]}
    \sigma(\mathbf{W}^{[2]}\mathbf{T}_{b_i}^\mathfrak{c}+\mathbf{b}^{[2]})
    +\mathbf{b}^{[3]})+\mathbf{b}^{[4]}),
\]
where $\mathcal{F}(\mathbf{T}_{b_i}^\mathfrak{c})\in\mathbb{R}^{D}$. 

Here, we have a sequence of detection and estimation DNNs analogous to the explanations provided for the Bayesian approach. For a detection DNN (d-DNN), the inputs are cloud liquid  (LWP) and ice (IWP) water paths, total columnar water vapor mass (WVP), CAPE, and 2-m air temperature. The output layer has three neurons $D=3$ to account for the target labels $\mathbf{t}_i(\mathbf{T}_{b_i})=(t_i^1,t_i^2,t_i^3)$, which is a one-hot vector that encodes near surface atmospheric conditions including ``no-precipitating", ``snowing", and ``raining". For example, $\mathbf{t}(\mathbf{T}_{b_i}^\mathfrak{c}) = [1,0,0]^T$ denotes a non-precipitating atmosphere near the surface.

For the estimation DNN (e-DNN) the inputs are $\mathbf{T}_{b_i}\in\mathbb{R}^n$ and surface precipitation rates associated with its $k$-nearest neighbors (i.e., $k=20$) in the training databases as described in Section \ref{Sec:B.4}. Feeding the DNN with $k$-nearest neighbor TBs help to localize its estimation. Clearly, the output layer has only one neuron $D=1$ to estimate the near surface precipitation rate. In this case, there are two different networks for estimating near surface rain and snowfall rates to avoid propagating estimation errors due to the use of different $Z$-$R$ and $Z$-$S$ relationships for rainfall and snowfall retrievals respectively.

\subsubsection{Activation and Cost Functions}

Clearly, the output of the network $\mathcal{F}(\mathbf{T}_{b_i}^\mathfrak{c})$ should remain in proximity of $\mathbf{t}(\mathbf{T}_{b_i}^\mathfrak{c})$ (detection) or $r_i$ (estimation), based on a distance metric. To that end, a {\it loss function} should be defined that needs to be minimized over all weights and biases of the network.

In precipitation detection, the network needs to provide an output of a three dimensional vector containing probabilities associated with each precipitation label. In particular, let the input to the last layer $L$ be $\mathbf{v}_i \in \mathbb{R}^D$. The $d^{\rm th}$ component $v_i^{d}$ shall be large when $\mathbf{T}_{b_i}^\mathfrak{c}$ belongs to $d^{\rm th}$ precipitation category. This intuition is encoded by the SoftMax operator
\[
\sigma_{\mathfrak{SM}}(v_i^d) = \dfrac{\exp{(v_i^d)}}
{\sum_{d=1}^{D} \exp{(v_i^d)}},
\]
that magnifies the large components of $\mathbf{v}_i$ and maps them onto a probability simplex with positive values that sum to unity.

In the last layer, the output of the SoftMax function should be as close as possible to unity in component $d$, when $\mathbf{T}_{b_i}^\mathfrak{c}$ belongs to label $d^{\rm th}$. Therefore, we need a cost function that minimizes the distance between the outputs of the network and one-hot vector of labels in the probability space. The categorical cross-entropy is a common cost function for this purpose 
\[
\mathcal{J}(\theta)=
-\dfrac{1}{M}\sum_{i=1}^{M} \left(\sum_{d=1}^{D}
t_i^d \ln\left({\sigma_{\mathfrak{SM}}\left[v_i^d(\theta)\right]}\right)\right),
\]
where $\theta$ is a vector representation of all unknown weights $\mathbf{W}^{[2:L]}$ and biases $\mathbf{b}^{[2:L]}$.

When it comes to e-DNN, all layer use a Rectified Linear Unit (ReLU) activation function
\[
    \sigma_{\mathfrak{RL}}(z_i)=\begin{cases}0 & z_i\leq0, \\
    z_i & z_i>0,\end{cases}
\]
where $z_i$ is the input to $i$th neuron in a specific layer.

The cost function for estimation of precipitation rate in e-DNNs is chosen as follows:
\[
\mathcal{J}(\theta)=
\dfrac{1}{M}\sum_{i=1}^{M}
{||\mathcal{F}(\theta)-r_i||}^p_p,
\]
where $r_i$ is observed surface precipitation rate, and ${\Vert\cdot\Vert}_p^p$ is the $p$-norm of a vector. For estimation of rainfall rate, we set $p=1$, while $p=2$ is considered for snowfall. The reason for departing from classic least-squares ($p=2$) for estimation of rainfall rate is due to heavier tail structure of rainfall than snowfall. The rainfall data have large extreme rates that can make the estimations prone to biases, when a least-squares loss is used.

\subsubsection{Training Algorithm}

In this paper, we use the Root Mean Squared Propagation (RMSProp) \citep{Hinton2012}, which is an extension to the Adaptive Gradient descent (AdaGrad) \citep{Duchi2011} algorithm to minimize the aforementioned cost functions. These algorithms both rely on the stochastic steepest gradient descent \citep{Robbins1951, Bottou2012} approach; however, the RMSProp modifies AdaGrad in a way that increases the chance of finding the global minimum, when the cost function is highly non-convex. In a general setting, the use of a steepest descent approach involves two main steps: ({\it i}) calculation of the expectation of the gradient vector over all training data points and ({\it ii)} a proper choice of learning rate or step size $\eta$ to assure a computationally efficient convergence.

As the size of the network and/or training data grows, the first step becomes computationally prohibitive and an optimal choice of $\eta$ becomes more important for a successful training. The stochastic gradient approach provides a cheap alternative by approximating the expected value of the gradients only over a much smaller number of randomly selected training data points. To that end, the training dataset is usually shuffled randomly and split into some mini-batches of size $M^{\prime} \ll M$.  Here, we use different batch sizes from 500 to 2000 for the detection and estimation networks through trial and error analyses.

\begin{figure}[t]
    \centering
    \includegraphics[width=0.50\textwidth]{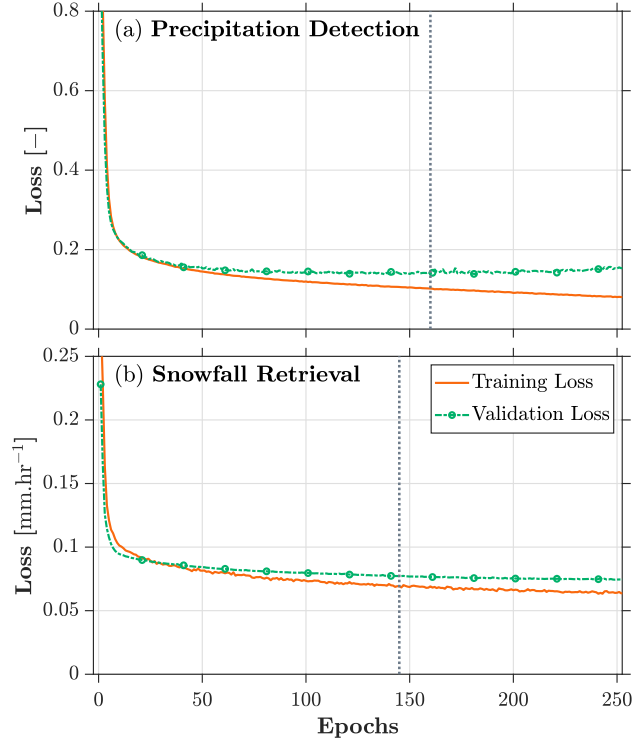}
    \caption{Examples of loss function changes as a function of training epochs for (a) detection (d-DNN) and (b) estimation DNNs (e-DNN) using DPR database over oceans. The initial learning rates are set to \num{1e-4} and \num{1e-5}, respectively. To avoid overfitting, the learning process stops at the point where the validation loss does not decrease in the next 25 epochs.}
    \label{fig:02}
\end{figure}

\subsubsection{Overfitting and Regularization} \label{Sec:B.4}

Note that only 1E+6 number of these coincidences over land and oceans are used for tuning the network parameters. To avoid overfitting, we split the tuning database of DPR and CPR coincidences into training (70\%), validation (15\%), and testing (15\%) data sets. The training dataset is used for learning the unknown parameters in the network. After each training epoch, validation data are fed to the trained network and the error is measured. Initially both training and validation errors decrease until there is no improvement in the performance of the network on the validation dataset at which the training is stopped. Two examples are provided in Fig~\ref{fig:02}. The testing dataset is used to evaluate the performance of the network on an independent data not used before in the learning process. To further boost generalization capability of the DNNs, we used the ``Dropout'' regularization \citep{Srivastava2014} that randomly ignores a fraction of neurons along with all its incoming and outgoing connections during the training process. Here, a dropout rate of 10\% is considered at each training epoch.

\subsubsection{Algorithmic Architecture}

\begin{figure*}
\centering \includegraphics[width=\textwidth]{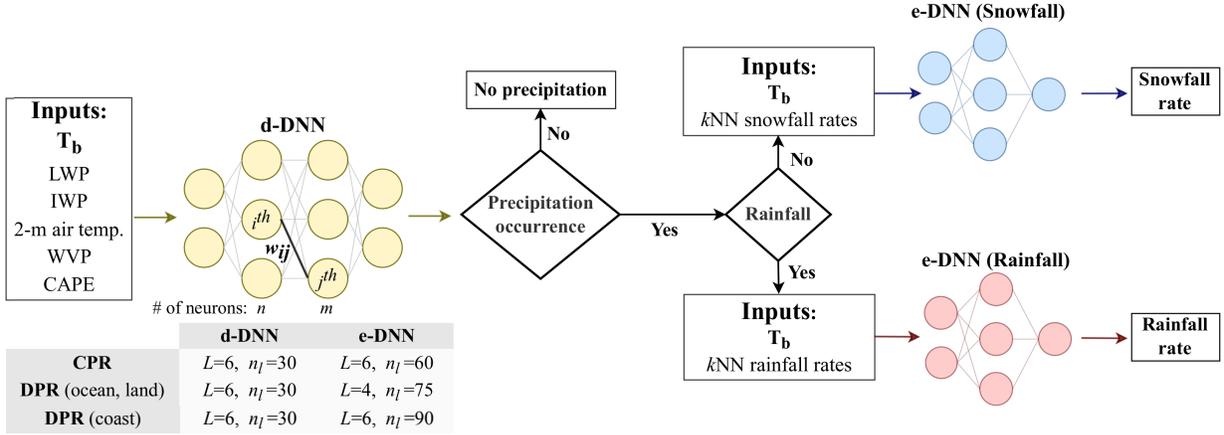}
    \caption{The architecture of the Deep-learning precIpitation rEtrieval alGOrithm (DIEGO) for a single {\it a priori} database over a specific surface type. As shown, a Deep Neural Network (DNN) first detects (d-DNN) the occurrence and phase of precipitation and then another one (e-DNN) estimates its rate. The inputs to the DNNs, number of layers $L$, and number neurons $n_l$ in each layer are shown for different training databases in the table.}
    \label{fig:03}
\end{figure*}

As explained previously, we propose a sequence of DNNs to first {\it detect} the surface precipitation occurrence and its phase and then {\it estimate} its near surface rate. This configuration is guided through the KerasTuner Python library \citep{OMalley2019} that trains a network with different configurations and hyperparameters (e.g., number of hidden layers, neural nodes) to find the best set of choices based on the defined cost function over a prespecified search space. The basic architecture of the algorithm is shown in Fig.~\ref{fig:03} for a specific surface type. As shown, for each surface type, we have two d-DNNs and four e-DNNs to detect precipitation and its phase and estimate its rate using both DPR and CPR observations.

\section{Results and Discussion} \label{sec:4}

The results of the algorithm are shown and discussed in three steps. First, we show the error metrics for the testing data sets. The results for orbital retrievals of a few selected storms including Winter Storm Grayson over the East Coast of the U.S., a snowstorm over Greenland, Hurricane Zeta over the Gulf of Mexico, and a storm over the Southern Indian Ocean. Lastly, we implement the algorithm for all GMI orbits in calendar year 2021 and compare its outputs with GPROF-V07 retrievals and ERA5 reanalysis data on an annual scale. 

In a storm-scale, we also present combined retrievals through a simple fusion approach only at a storm-scale. When either of CPR or DPR d-DNNs detects precipitation, we label that pixel as precipitating. When the detected phase is different, the pixel is labeled as mixed. It is important to note that, this does not literally imply a mixed phase precipitation and just represents a higher degree of retrieval uncertainty in terms of the detected phase by the two active databases. When only one of the d-DNNs detects precipitation, we retrieve its rate solely based on the sequenced e-DNN and use the results as the final retrieval. When both d-DNNs for CPR and DPR detect precipitation, the final rate is considered to be the mean of the estimated rates. 

\subsection{Retrieval experiments using the test data sets}

The test data sets are used to quantify intrinsic uncertainties in terms of detection and conditional estimation accuracy. The detection accuracy is quantified based on the True Positive Rate (TPR) or probability of detection and False Positive Rate (FPR) or probability of false alarm in Table.~\ref{tab:01}. In the table, we also report the results of $k$NN classification with $k=20$, based on a majority vote rule, which was shown to be extremely effective in the PWM detection of precipitation and its phase \citep{EbtBF15,Ebtehaj_2016_eval,EbtKum17,Vahedizade2021}.

\begin{table}[h]
\caption{Detection performance of DIEGO and $k$NN with $k=20$. True Positive Rate (TPR) and False Positive Rate (FPR) are reported for retrievals based on both CloudSat CPR and GPM DPR testing databases.}
    \scriptsize \centering \renewcommand{\arraystretch}{1.1}
\begin{tabular}{|ccccccc|c|}
\cline{2-7} \cline{3-7} \cline{4-7} \cline{5-7} \cline{6-7} \cline{7-7} 
\multicolumn{1}{c|}{} & \multicolumn{6}{c|}{{\bf CPR}} & \multicolumn{1}{c}{}\tabularnewline
\cline{2-7} \cline{3-7} \cline{4-7} \cline{5-7} \cline{6-7} \cline{7-7} 
\multicolumn{1}{c|}{} & \multicolumn{2}{c|}{{\bf Ocean}} & \multicolumn{2}{c|}{{\bf Land}} & \multicolumn{2}{c|}{{\bf Coast}} & \multicolumn{1}{c}{}\tabularnewline
\cline{2-7} \cline{3-7} \cline{4-7} \cline{5-7} \cline{6-7} \cline{7-7} 
\multicolumn{1}{c|}{} & \multicolumn{1}{c|}{TPR} & \multicolumn{1}{c|}{FPR} & \multicolumn{1}{c|}{TPR} & \multicolumn{1}{c|}{FPR} & \multicolumn{1}{c|}{TPR} & FPR & \multicolumn{1}{c}{}\tabularnewline
\hline 
{\bf $k$NN} & 89.0 & 2.3 & 99.9 & 0.2 & 88.5 & 2.1 & \multirow{2}{*}{{\bf Rain}}\tabularnewline
{\bf DIEGO} & 92.3 & 3.0 & 99.9 & 0.1 & 94.8 & 1.7 & \tabularnewline
\hline 
{\bf $k$NN} & 95.4 & 4.4 & 92.3 & 6.1 & 95.2 & 5.7 & \multirow{2}{*}{{\bf Snow}}\tabularnewline
{\bf DIEGO}  & 96.6 & 2.5 & 95.5 & 3.1 & 98.9 & 1.9 & \tabularnewline
\hline 
\multicolumn{1}{c|}{} & \multicolumn{6}{c|}{{\bf DPR}} & \multicolumn{1}{c}{}\tabularnewline
\hline 
{\bf $k$NN} & 84.8 & 3.9 & 86.8 & 2.6 & 78.7 & 3.6 & \multirow{2}{*}{{\bf Rain}}\tabularnewline
{\bf DIEGO}  & 96.0 & 7.0 & 97.2 & 3.7 & 96.1 & 8.2 & \tabularnewline
\hline 
{\bf $k$NN} & 83.2 & 4.9 & 83.0 & 6.9 & 88.0 & 6.8 & \multirow{2}{*}{{\bf Snow}}\tabularnewline
{\bf DIEGO}  & 93.3 & 3.5 & 96.2 & 2.7 & 91.7 & 3.9 & \tabularnewline
\hline 
\end{tabular} \label{tab:01}
\end{table}

\begin{figure*}
    \centering
    \includegraphics[width=\textwidth]{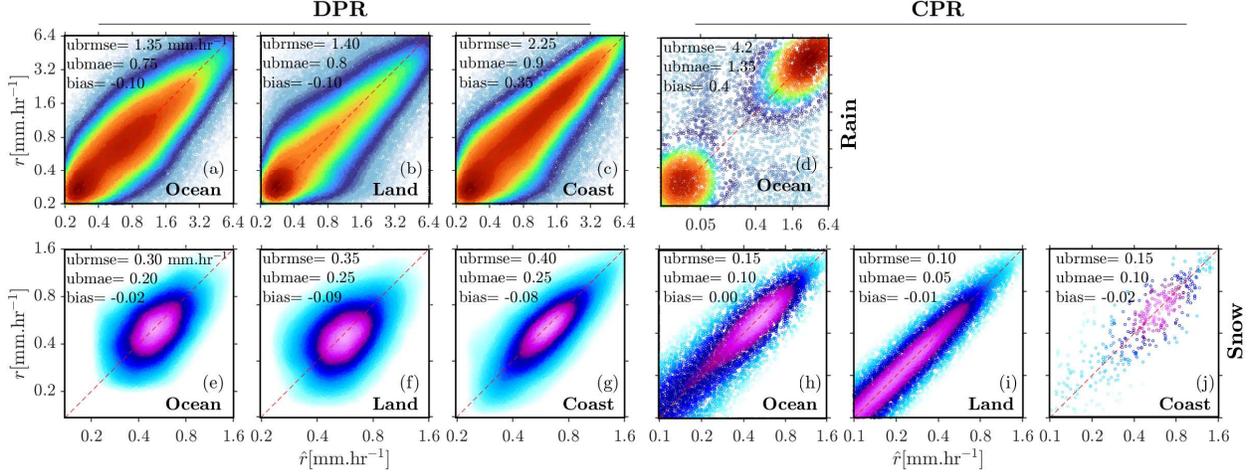}
    \caption{Passive microwave retrievals of rainfall (a--d) and snowfall (e--j) by the DIEGO e-DNNs using the test databases for GPM DPR and CloudSat CPR. Variation of color from red--blue (blue--magenta) shows the variation of density of points from 0--1 in rainfall (snowfall) retrieval plots. from  The quality metrics are bias, unbiased root mean squared (ubrmse) and unbiased mean squared error (ubmae).}
    \label{fig:04}
\end{figure*}

The results show that the d-DNNs, with intrinsic classification capabilities as well as inclusion of relevant physical variables, generally provide a superior performance compared to $k$NN in detection of precipitation and its phase. Note that matching through $k$NN is conducted only in the TB space. The probability of rainfall (snowfall) detection over oceans, using CPR data, reaches to almost 92 (97)\%, while the probability of false alarm is below 3.0 (2.5)\%. The CPR snowfall retrievals over land exhibit a TPR of $\approx$ 95\% with the false alarm probability of less than 4\%. For the DPR rainfall (snowfall), the TPR is above 96\% (92\%) over all land surfaces and is maximum over land 97.2 (96.2)\%. The probability of false alarm is consistently below 8.0\%, which is obtained for coastal rainfall. The FPR in detection of snowfall is slightly lower than rainfall using DPR databases and remains below 4\% over land and oceans. Overall, we observe that the metrics are superior for rainfall (snowfall) retrievals when using the DPR (CPR) database.

The results of DIEGO e-DNNs are presented in Fig.~\ref{fig:04} for both rainfall (top row) and snowfall (bottom row) conditional retrievals using DPR (first three columns) and CPR (last three columns) test databases -- given that the precipitation is detected and its phase is properly determined. The statistics of the intrinsic uncertainties are reported for active retrievals below the 97.5th percentile to alleviate the effects of a few extreme values. As previously explained, the estimation results are not provided for CPR rainfall over land and coastal zones. It is important to note that the scatter plots are shown after applying a cumulative density function (CDF) matching to remove the reported biases. The CDF matching functionals were learned from the testing databases for both CPR and DPR data. The most important observation is that PMW DPR rainfall retrievals are less uncertain than CPR rainfall over oceans. However, CPR snowfall retrievals exhibit less uncertainty than DPR snowfall, which also corroborate with the reported detection metrics in Table \ref{tab:01}.   

As reported in Fig.~\ref{fig:04}, the bias is relatively under control and is larger for rainfall than snowfall. Generally speaking, when the probability density function of the data is positively skewed, the estimators, which tend to regress towards the mean, lead to negative biases and vice versa. For example, the bias is negative in DPR precipitation retrievals over land and oceans except over coastal areas, which is  around 0.35~\si{mm.hr^{-1}}. Therefore, it appears that in these case, there are extremely light rainfall values that populate the testing data, which can be overestimated. However, the largest negative biases are over land, indicating the presence of more frequent extreme convective rainfalls events. The biases in snowfall retrievals are less than 0.1~\si{mm.hr^{-1}} and slightly negative in majority of land surface types for both CPR and DPR -- indicating that heavy snowfall events are underestimated; however, to a lesser extent than rainfall extremes.

The uncertainty of retrievals in terms of unbiased root mean squared error (ubrmse) and unbiased mean absolute error (ubmae) provides important information about the capability of the algorithm in retrieval of precipitation over different surface types. We can see that these uncertainty metrics are larger for rainfall than snowfall, which is expected as the probability distribution of rainfall is more heavy tailed. Overall, we observe that DPR (CPR) snowfall ubrmse is around 25 (4)\% of the rainfall ubrmse. The maximum ubrmse for rainfall and snowfall belongs to CPR over ocean (4.2 \si{mm.hr^{-1}}) and DPR over coast (0.4\si{mm.hr^{-1}}), respectively. The minimum ubrmse of 0.10 \si{mm.hr^{-1}} is obtained for CPR snowfall retrievals over land. Once again, these observations indicate that intrinsically, DPR data lead to reduced uncertainty in retrieval of rainfall compared to CPR data, while this pattern is reverse when it comes to snowfall retrievals.  

\begin{figure*}
    \centering
    \includegraphics[width=0.5\textwidth]{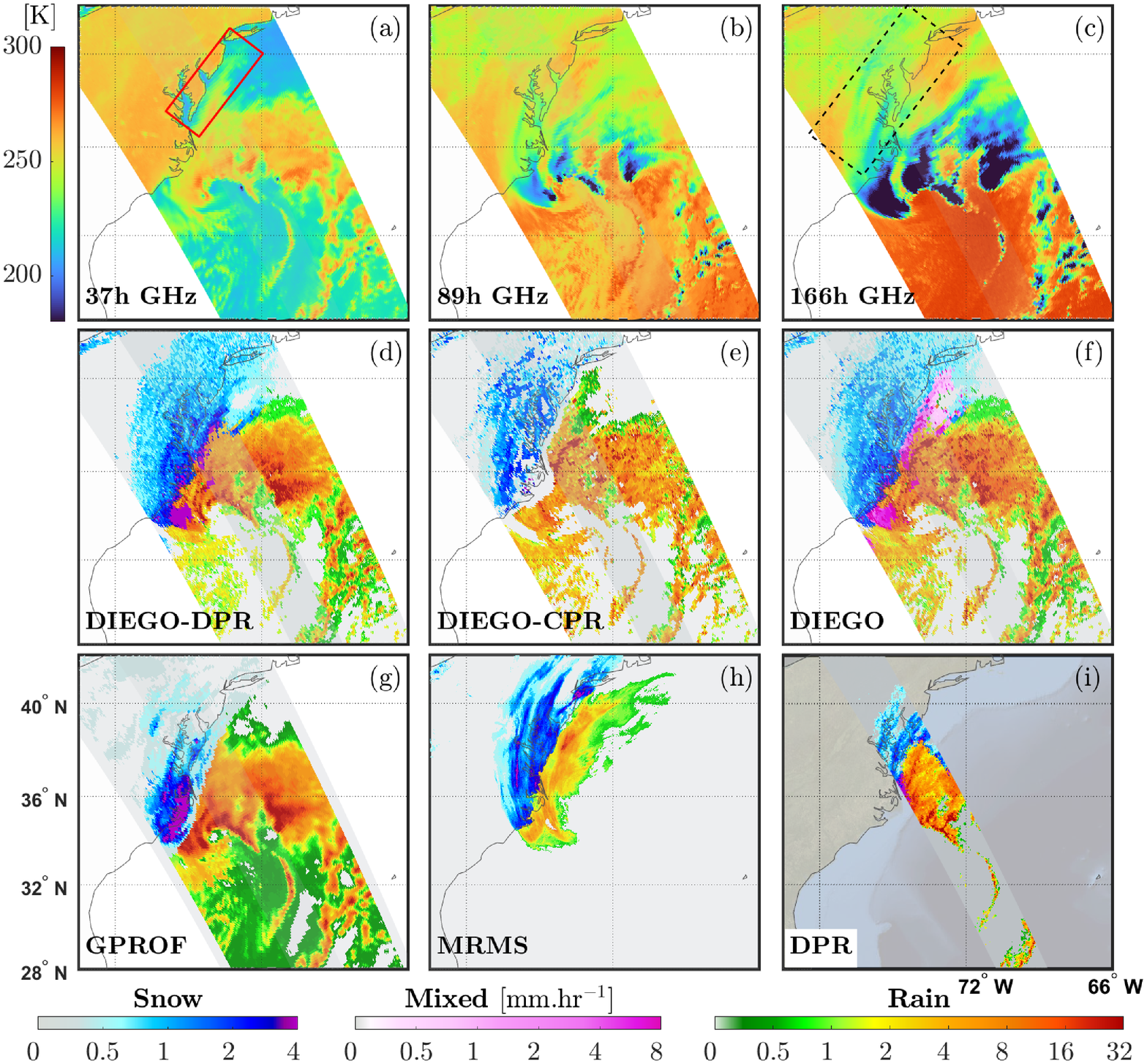}\includegraphics[width=0.5\textwidth]{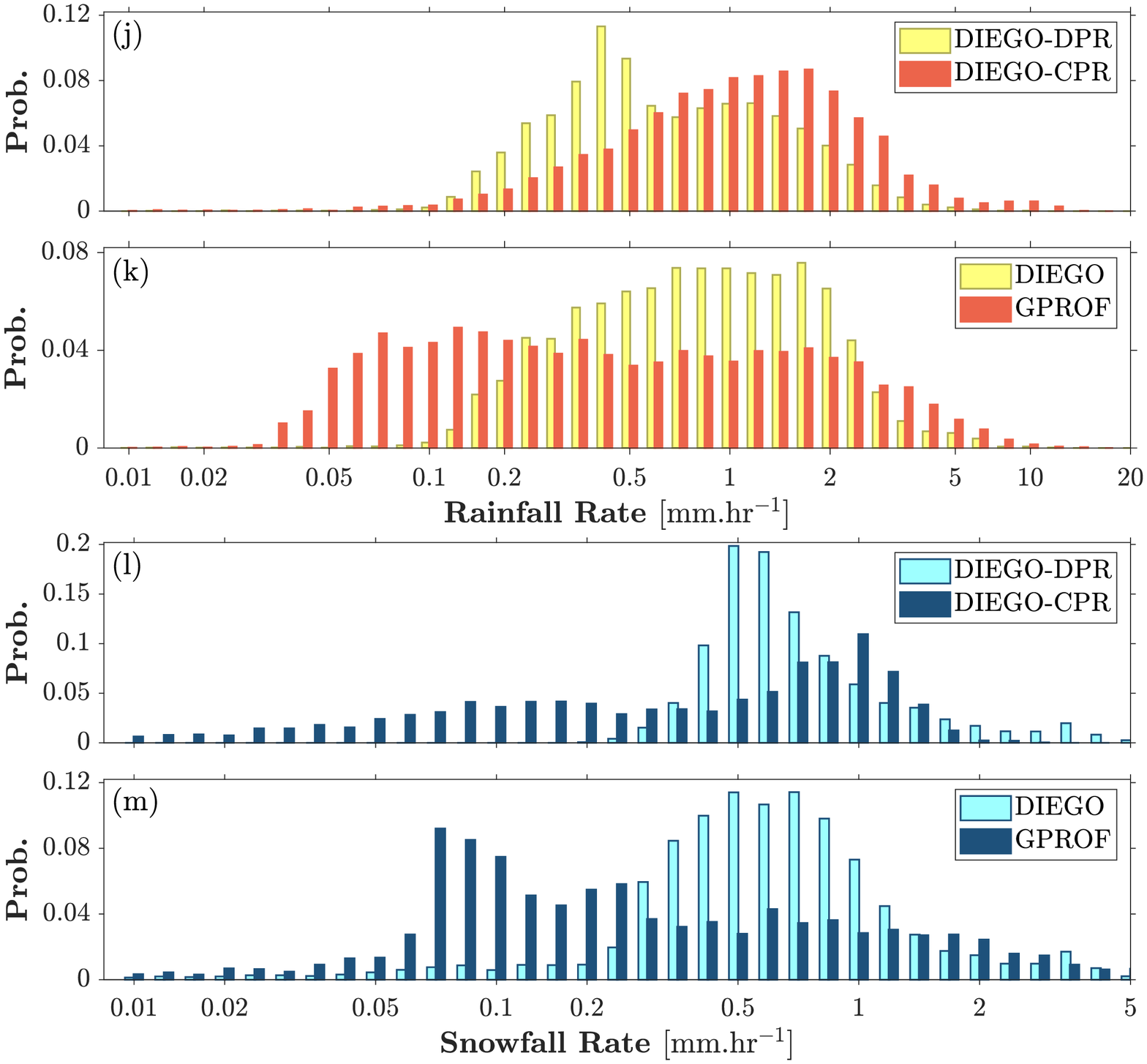}
        \caption{Passive microwave retrievals (GPM orbit \#21882) capturing the precipitation dumped by the Winter Storm Grayson on January 4, 2018 over the East Coasts of the United States. The GMI TBs at 37, 89 and 166 GHz horizontal polarization (a--c), precipitation retrievals by DIEGO (d--f), GPROF, MRMS, and DPR (g--i), as well as their corresponding probability histograms (j--m) at different phases.}
    \label{fig:05}
\end{figure*}

\subsection{Storm-scale retrieval experiments}

Fig.~\ref{fig:05} shows the results of the DIEGO applied on a GMI overpass (orbit \#21882) capturing the powerful North American blizzard, known as the Winter Storm Grayson, that caused widespread severe disruption and loss of life across the East Coast of the United States and Canada in early January 2018. Figs.~\ref{fig:05}a--c show the horizontally polarized TBs at 37, 89 and 166 GHz. Warmer signals of raindrop emission are apparent in 37 GHz over radiometrically cold Atlantic ocean. At higher frequencies, significant scattering signal or TB cooling is observed -- indicating high concentration of ice aloft over ocean and the presence of a melting layer.  

\begin{figure*}[t]
    \centering
    \includegraphics[width=.85\textwidth]{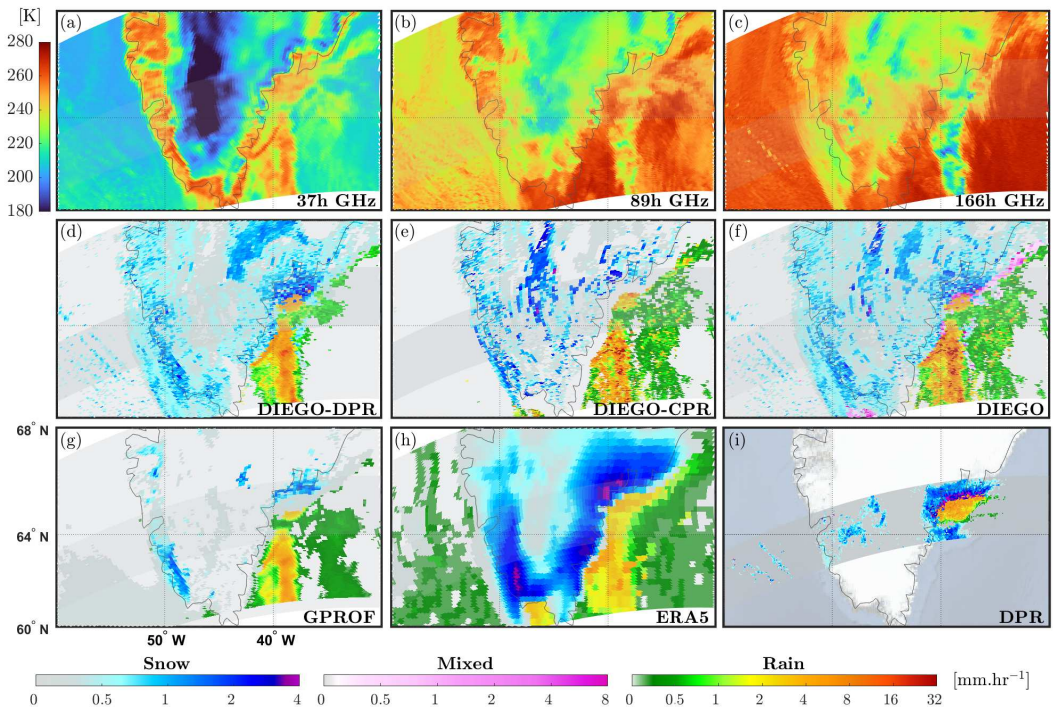}
    \caption{Brightness temperatures by GMI (a--c), precipitation retrievals by DIEGO (d--f) and those from GPROF, ERA5, and DPR (G--I) for orbit \#3080 on September 13, 2014 over Greenland.}
    \label{fig:06}
\end{figure*}

The results of DIEGO retrievals based on the DPR and CPR databases (Fig.~\ref{fig:05}d,e) are relatively consistent over ocean in terms of detected precipitation phase. The storm bands are properly captured in both retrievals. Even though, the DIEGO-CPR mean rainfall overestimates its DPR counterpart (Fig.~\ref{fig:05}j,k). Unlike DPR-based retrievals, those using the CPR database, show less variability as the rainfall rate remains almost constant across the central part of the storm and its trailing bands. For the retrieved rainfall values, the 97.5th percentile in DIEGO-CPR (-DPR) is 16.5 (9) \si{mm.hr^{-1}}; however, the 99.5th percentile is 38 (60) \si{mm.hr^{-1}}. In other words, while the bulk of the CPR-based retrievals overestimate the DPR counterpart, the extreme cells are more likely in DIEGO-DPR. We can clearly see that the GPROF retrieves a wider range of rainfall variability with a flatter probability distribution and a large probability mass below 0.1~\si{mm.hr^{-1}}, than DIEGO retrievals.  

In the Delaware coastlines region, we clearly see that the warming signal in 37 GHz wanes down (shown within a red box), indicating that the rainfall is transitioning to snowfall from ocean to land. Visual inspection of 89 GHz TBs, shows that an outer band of depressed TB values is observed extending from southeast Chesapeake Bay to the Bald Head Island in North Carolina. This radiometrically cold band extends to northeast Pennsylvania in 166 GHz (shown within a black box), indicating the probability of overland snowfall. A large radiometrically cold area also exists in the 37 to 166 GHz observations over the northwest sector of the overpass over Pennsylvania. This features are likely due to ground snow cover and not precipitating snow. 

All shown passive and active precipitation retrievals as well as MRMS (Fig.~\ref{fig:05}h) observations consistently indicate that the precipitation phase changes from liquid to solid as the storm moves from over ocean to land. The extent of snowfall occurrence is relatively consistent between DIEGO-DPR and -CPR. The DIEGO-CPR retrievals are less coherent and patchier than DPR, which can be due to the fact that GMI-CPR coincidences included in the database are $\pm$15 minutes apart, which might lead to noisier CPR retrievals. As is evident, over the outer bands of the storm, the DIEGO-CPR retrieves lighter snowfall than DIEGO-DPR, which is more consistent with the MRMS data. A combined version of the retrievals is presented in (Fig.~\ref{fig:05}f). As explained previously, when CPR and DPR are inconsistent with respect to the detected phase, the retrievals are labeled as {\it mixed} and shown with a pink color map. As is evident, those mixed retrievals are mostly along the coastlines, where precipitation is changing its phase.   

The probability histograms of DIEGO-DPR and -CPR snowfall retrievals, their combined version (DIEGO), as well as GPROF retrievals are shown in Fig.~\ref{fig:05}j--m. The mean snowfall in DIEGO-DPR is $\sim$0.85 \si{mm.hr^{-1}} while it is around 0.5 \si{mm.hr^{-1}} in DIEGO-CPR and thus, unlike the rainfall, snowfall is overestimated using DPR database. We can see the distribution of DIEGO-CPR is much wider than DIEGO-DPR and extends from 0.01 to 2 \si{mm.hr^{-1}}. The 97.5th percentiles are 3.5 and 1.5 \si{mm.hr^{-1}} and maximum retrieved values are 7.5 and 3 \si{mm.hr^{-1}}, for DIEGO-DPR and -CPR, respectively. Visual inspection of the histograms show that GPROF snowfall rates are largely concentrated below 0.1 \si{mm.hr^{-1}}, while a few extreme values exceeding 3 \si{mm.hr^{-1}} thicken the tail of the distribution significantly. These extreme values are largely extended from south of the Chesapeake Bay to the Pamlico Sound lagoon in North Carolina -- shown with magenta color in Fig.~\ref{fig:05}g.  

A complementary snowfall event at higher latitudes was also analyzed using GMI observations and various satellite-based retrievals. Fig.~\ref{fig:06} shows a snowstorm over Greenland on September 13, 2014 (GMI orbit \#3080). Passive microwave retrieval of precipitation over Greenland ice sheet can be challenging because of two main reasons. First, the background surface emission from a snow-covered ice sheet is different than majority of land surfaces. Thus, statistically speaking, there exist less coincidences of active/passive observations over this specific surface type and naturally the learned relationship between the TBs and surface precipitation can be more uncertain. Second, over a less emissive snow-covered ice, snowfall high-frequency scattering signatures can be weaker and the masking effects of supercooled cloud liquid water content can be stronger. In other words, under similar atmospheric conditions, the expected high frequency TB cooling for snowfall events over Greenland might be less prominent compared to cases when snow falls over more emissive surfaces such as those covered by a vegetation canopy.

Visual inspection of 37 GHz channel (Fig.~\ref{fig:06}a) indicates increased TB values and thus potential occurrence of a rainfall events over the North Atlantic Ocean and Greenland Strait. This warm signal is accompanied with TB cooling in both 89 and 166 GHz -- indicating ice aloft scattering and thus can show the presence of a melting layer. Over the southern and western coastlines, we observe a warming signal at 37 GHz, largely over ice free areas with a sub-arctic and tundra climate regimes. Perceptual interpretation of the TBs at high frequencies in response to the occurrence of precipitation is not straightforward due to the complexity of radiometric interactions of snow-covered ice with atmospheric signals -- especially at 89 GHz, which is more sensitive to changes of surface emission than 166 GHz. Nevertheless, a cooling signal, likely due to the presence of ice particles and falling snow, is clearly visible over the southwest coastlines.   

Although, it appears that as the storm moves over the ice sheet, the phase of precipitation transitions from liquid to solid, the snowfall retrievals are notably different across the shown products. The ERA5 simulations (Fig.~\ref{fig:06}h) indicate the occurrence of a snowstorm over southern Greenland along the coastlines. The intensity reaches to almost 4~\si{mm.hr^{-1}} over the coastal areas and decays to zero in the middle of the ice sheet. While DIEGO retrievals also capture a coastal snowstorm, the retrieved rates are lower than ERA and some patchy snowfall cells are detected in the middle of ice sheet (Fig.~\ref{fig:06}d--f), which are not consistent between the DPR and CPR based retrievals. We suspect that the DPR database provides a more skillful detection capability over radiometrically complex surfaces due to higher number of samples of such conditions in its database. However, as shown in Fig.~\ref{fig:01}, the DPR-based snowfall retrievals are generally prone to overestimation compared to those of CPR.  

\begin{figure*}
    \centering
    \includegraphics[width=\textwidth]{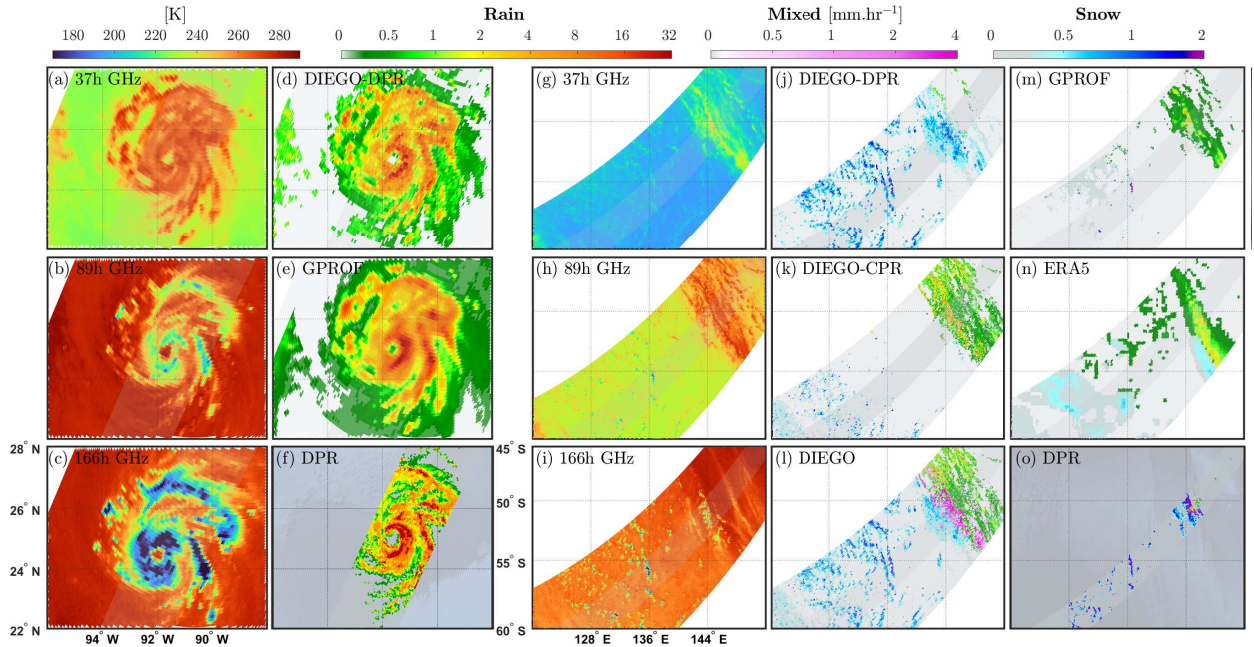}
    \caption{Precipitation retrievals using the observed brightness temperatures for orbit \#37875 on October 28, 2020 (Hurricane Zeta) and orbit \#3498 on October 10, 2014.}
    \label{fig:07}
\end{figure*}

To further demonstrate the performance of the proposed algorithm, precipitation retrievals for two additional orbits on October 10, 2014 (\#3498) and October 28, 2020 (\#37875) are shown in Fig.~\ref{fig:07}. Orbit \#37875 captured the Hurricane Zeta that made landfall over Louisiana, United States. Overall, there is consistency between all shown retrievals. Here, we only show DIEGO-DPR because we have observed that the retrievals for intense rainstorms are more realistic using DPR than the CPR database as CPR reflectivity saturates. As is evident, DIEGO-DPR captures the intensity of rainfall and its spatial band structure in a similar fashion as to GPROF and original DPR observations. As expected the intensities are generally underestimated for extreme values while the extent of lighter rainfall is overestimated compared to the active counterpart.  

The results for orbit \#3498 capturing a storm over the southwest Indian Ocean are different among the shown retrievals. Over the northeast part of the storm, 37 GHz observations show a warming signal indicating the possibility of a frontal precipitation event. Over higher frequency channels, there is no signature of ice scattering, except over the southern edge of the front -- indicating that the clouds are largely warm, lifting is shallow, and potential precipitation is likely to be in liquid phase. The majority of retrievals indicate liquid precipitation over this part of the storm, including GPROF, DIEGO-CPR and ERA5, except the active DPR and passive DIEGO-DPR retrievals. Active DPR does not detect any rainfall and only catches snowfall events. Therefore, we conjecture that the observed discrepancies are not due to an algorithmic flaw but are rather related to intrinsic characteristics of the DPR training database. The DPR active retrievals use the reanalysis data of air temperature and moisture from the operational global analysis provided by the Japan Meteorological Agency, while CPR active retrievals rely on the ERA5 reanalysis data and use two different approach for phase determination. Even though, we fed the d-DNNs with ERA5 2-m air temperatures, it appears that the connections between TBs and precipitation phases learned from the DPR database significantly influence the final e-DNN outputs. 

Over the southwest part of the orbit, weak sporadic warm signals exist at 37 GHz, which can be due to the presence of supercooled liquid water in the clouds. However, unlike the northern part of the storm, some of these warm signatures contain convective cold cells capturing significant high-frequency scattering signatures, perhaps due to the presence of localized convection. The shape of the signatures indicates possibilities of postfrontal cumuliform type clouds that contribute $\sim 35$\% of global snowfall events \citep{Kulie2016}. Sporadic snowfall is detected over those cells in all retrievals with different intensity values and extent. DIEGO-DPR retrieves more sporadic snowfall events over those high-frequency depressions with higher rates than the DIEGO-CPR, which is consistent with previous observations.

\begin{figure*}
    \centering
    \includegraphics[width=\textwidth]{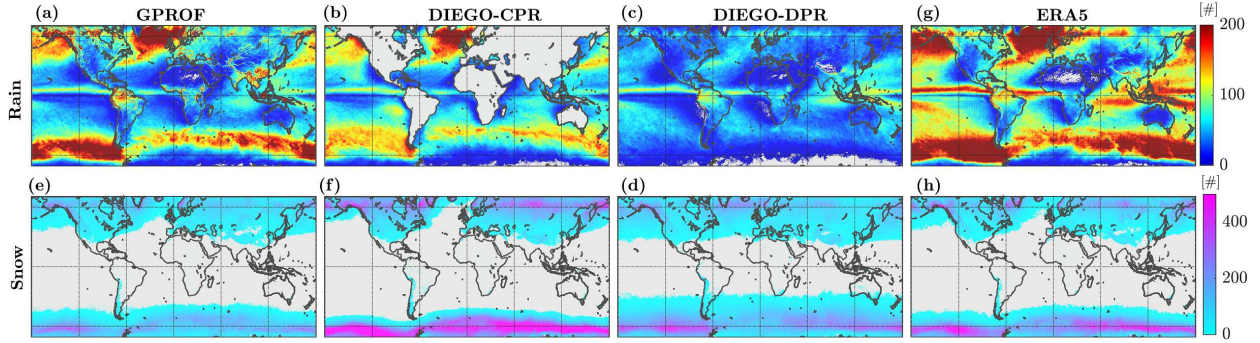}
    \caption{Annual occurrence number of rainfall (top row) and snowfall (bottom row) in 2021 mapped onto a 0.1$^\circ$ grid box for GPROF (a,\,e), DNN-CPR (b,\,f), DIEGO-DPR (c,\,g), and ERA5 (d,\,h) products.}
    \label{fig:08}
\end{figure*}

\subsection{Global-scale retrieval experiments}

Throughout this subsection, we will compare the annual ERA5 data with the satellite retrievals. We should emphasize that ERA5 data should not be considered as a ground-truth; however, since the conservation of water and energy is satisfied in an Earth system model, its annual mean values might be considered as a relative reference for consistency assessment of satellite retrievals.

Figs.~\ref{fig:08}a--h show the annual frequency of rain and snowfall detection in 2021 for the rates above 0.01 \si{mm.hr^{-1}} --- considering all overpasses of GMI. Note that the ERA5 data, with a resolution of 30~\si{km}, are mapped onto the GMI grids using linear (nearest neighbor) interpolation in time (space). We did not retrieve any precipitation over sea ice as the number of DPR and CPR coincidences were not adequate for a robust training of the neural networks. Therefore, for a one-to-one comparison, we removed precipitation data from all products over sea ice.

As is evident, there are some discrepancies between the frequency of occurrences among the shown annual retrievals -- especially for the rainfall data. The differences are expected, partly due to two main reasons. First, ERA5 cannot properly represent sub-grid scale convective precipitation processes. This leads to a representative error, and a positive bias in occurrences, especially for rainfall (Fig.~\ref{fig:08}a--d). Because any sub-grid occurrence of precipitation shall be represented with a lower rate across the entire ERA5 grids, to warrant an unbiased estimation. Beyond the middle latitudes, we still observe that ERA5 has higher frequency of occurrence over the Intertropical Convergence Zone (ITCZ) than all other satellite products, which further verify the provided reasoning. Second, precipitation radars cannot properly capture light precipitation, especially DPR retrievals that are limited to the rates greater than 0.2~\si{mm.hr^{-1}}. As a result, large discrepancies are apparent over mid-latitude oceans, where the rainfall events are light but frequent. For example, we observe a zonal belt of high occurrences, in all products expect DIEGO-DPR, over the south Pacific ocean around latitude 50$^\circ$S that extends to the southern Atlantic and Indian oceans over latitudes 30--40$^\circ$S.

\begin{figure*}[t]
    \centering
    \includegraphics[width=\textwidth]{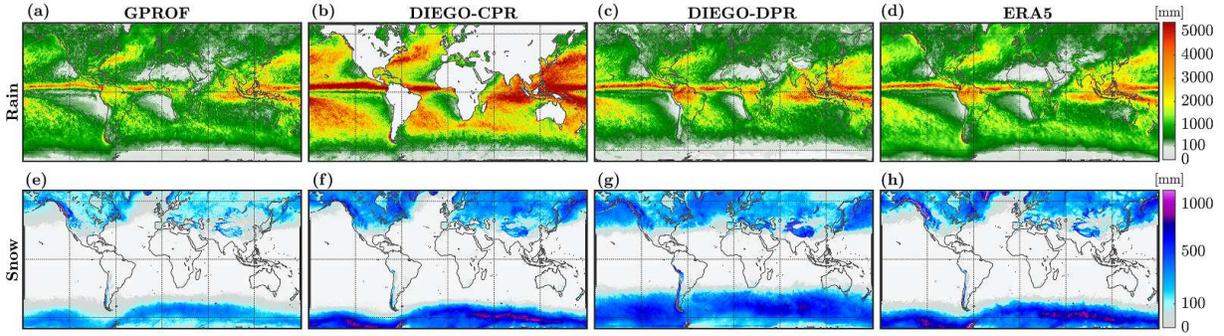}
    \caption{Annual estimates of the total rainfall (top row) and snowfall (bottom row) (in millimeters) in 2021 at a 0.1$^\circ$ grid box for GPROF, DIEGO-CPR, DIEGO-DPR, and ERA5 products.}
    \label{fig:09}
\end{figure*}

Overall, for rainfall retrievals (Fig.~\ref{fig:08}a--d), we see that the occurrence frequency in DIEGO-DPR is less than 50\% of ERA5 and GPROF in middle latitudes. We need to recall that the true positive rate of rainfall detection for DIEGO-DPR is more than 95\% over land and oceans for controlled retrieval experiments (Tab.\ref{tab:01}). Therefore, passive retrievals based on DPR data can miss significant number of rainfall events. We need to note that detection deficiencies may not necessarily lead to significant underestimation in the total amount as shown later on in Fig.~\ref{fig:09}c. DPR retrievals often capture intense convective precipitation at a much higher resolution than the reanalysis data, which can compensate for missing light rainfall when calculating total annual precipitation. The occurrence rates significantly increase for DIEGO-CPR as the database contains much lighter precipitation rates than the DPR. This indicates that even though the rainfall CPR database may not lead to unbiased rainfall retrievals with an expected dynamic range, it provides higher detection accuracy of light rainfall compared to the DPR database.

For snowfall (Fig.\,\ref{fig:08}e--h), DIEGO-CPR provides the largest number of occurrences, which are apparent over the polar regions. The number and extent of occurrences are relatively consistent between DIEGO-CPR and ERA5 -- perhaps because snowfall is less driven by intense convection and its dynamic range is narrower than rainfall. The lowest occurrence rates belong to DIEGO-DPR, however, the snowfall occurrences extend to lower latitudes and for example cover parts of the Southern Australia. Once again, as demonstrated in Fig.~\ref{fig:09}, small number of occurrences may not directly translate into a lesser amount of the total annual snowfall. 

\begin{figure}
    \centering
    \includegraphics[width=0.50\textwidth]{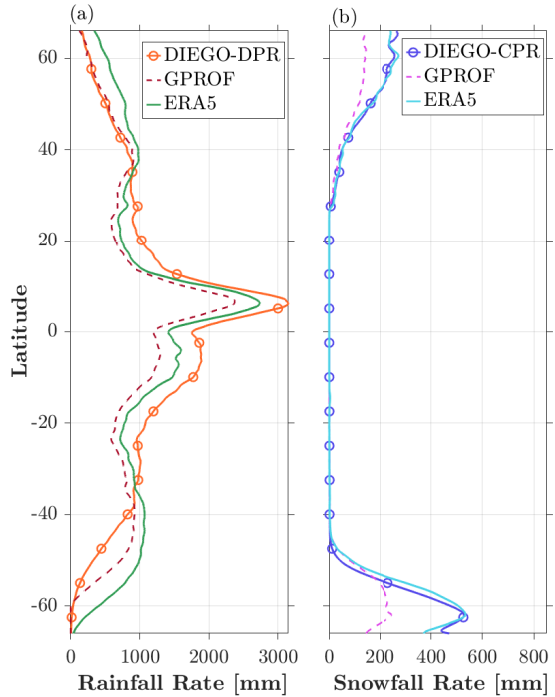}
    \caption{Annual rainfall (a) and snowfall (b) zonal mean values obtained from Fig.~\ref{fig:09}.}
    \label{fig:10}
\end{figure}

An estimate of total annual precipitation for all products is shown in Fig.~\ref{fig:09}. For the rainfall retrievals (Fig.~\ref{fig:09}a--d) there is consistency among all products except DIEGO-CPR, which significantly overestimates the global rainfall as previously discussed. Visual inspection shows that DIEGO-DPR slightly overestimates ERA5 and GPROF over the ITCZ where majority of rainfall events are convective; however, underestimates those products over lower latitudes where stratiform light rainfall can dominate precipitation regimes. This latitudinal trend is more clear in Fig.~\ref{fig:10} that shows zonal mean of precipitation for a subset of selected products. As shown, above 35$^\circ$ S-N, DIEGO-DPR underestimates both ERA5 and GPROF and overestimate them within the tropics.

The results of annual snowfall (Fig.~\ref{fig:09}e--h) show marked differences between the products both over land and oceans. Overall, it appears that DIEGO-DPR (GPROF) overestimates (underestimate) snowfall both over land and oceans, compared to other products. There is a reasonable degree of agreement between DIEGO-CPR and ERA5 data both over oceans and land. For example, over Eurasia and Tibetan Plateau, the annual retrievals both in terms of spatial patters and magnitude are closely similar while GPROF significantly underestimate both ERA5 and DIEGO-CPR. The snow depth climatology of Eurasia \citep{Bormann2018,He_2018} indicates that on average we should expect 50~\si{cm} of snow over Siberia, which is consistent with DIEGO-CPR. 

However, over some areas covered with shallow seasonal snow-cover, such as the Canadian Prairie, we observed that DIEGO retrievals, for both CPR and DPR databases, overestimate other products. Recent research \citep{Rahmi2022} shows that over shallow snow cover with minimal emerging vegetation, where the depth is less than 10--15 \si{cm}, the PMW retrieval of snowfall can be excessively sensitive to change of snow layering and grain size distribution. Therefore, snow cover continues to be a major challenge for snowfall retrievals over land and sea ice and stratification of the database, based on the snow types and physical characteristics, can lead to reduced uncertainties even when a deep neural network is deployed. Nevertheless, the zonal mean of snowfall retrievals (Fig. \ref{fig:10}\,b) also indicates that DIEGO-CPR and ERA5 retrievals are in a close agreement. Over the Northern Hemisphere, we suspect that the overestimation by DIEGO-CPR is largely related to those snowfall events over snow-covered surfaces.

We need to emphasize that in Fig.~\ref{fig:09}, the retrieval of DIEGO-DPR snowfall retrievals are biased corrected. In \ref{sec:appedix}, the positively biased retrievals are shown and compared with active DPR data over the GPM inner swath (see Fig.~\ref{fig:1-Apx} in \ref{sec:appedix}). This overestimation can be caused by multiple reasons: i) The distribution of snowfall in DPR dataset is not heavy tailed with its mean value significantly higher than snowfall in CPR dataset. ii) The phase variable of 2A-DPR dataset is determined at the lowest radar range gate uncontaminated by surface clutters, which may be 0.5--2.0 \si{km} above the surface (even over oceans). Therefore, the actual phase might be different those retrieved more near the surface. iii) 2A-DPR detects more snowfall over ocean in the northern hemisphere compared to the other products, especially over the Norwegian and Greenland Sea. These snowfall values paired with their observed TBs are included in the training dataset and can affect the detection and estimation performance of DIEGO-DPR. To remove the existing bias, the snowfall retrieval of DIEGO-DPR are modified using a scale factor by dividing the latitudinal mean of active DPR snowfall retrievals by latitudinal mean of passive DPR snowfall retrievals, which can also be applied to the orbital retrievals as well.

\section{Summary and Conclusion} \label{sec:5}

In this paper, we examined a deep and dense neural network architecture that learns from coincidences of passive/active observations from the Global Precipitation Measurement (GPM) core satellite microwave imager (GMI), the W-band {\it CloudSat} Profiling Radar (CPR), and the Dual-frequency Precipitation Radar (DPR) onboard GPM. The algorithm can properly condition the retrievals to key cloud microphysical and environmental variables that are tightly linked to the occurrence, phase and rate of precipitation including the cloud liquid and ice water path, total water vapor content, 2-m air temperature, and convective potential energy -- obtained from ERA5 reanalysis data. This architecture first detects the occurrence of precipitation, determines its phase, and then estimates its rate through a series of two deep neural networks. 

Overall, under controlled numerical experiments, reported in Table \ref{tab:01}, we observed that DIEGO can improve detection of precipitation and its phase beyond previous Bayesian retrieval algorithms that have used {\it k}-nearest neighbor ({\it k}NN) matching techniques and neural network based approaches. However, the tested deep neural networks were unable to properly estimate the precipitation rate with an acceptable uncertainty -- solely from a pixel-level observed vector of brightness temperatures and other ancillary inputs. To reduce the uncertainty, the networks were fed with precipitation rates associated with {\it k}-nearest vectors of brightness temperatures in the training databases.  

Overall, evaluating the retrievals for multiple storms and assessing their annual values in 2021, the following key points are worth noting. (i) It is recommended that the coincident CPR snowfall data will be used in any future attempts for improving official algorithms or in new snowfall data products as recommended in other previous research as well \citep{Panegrossi2017,Ebtehaj2020,Sano2022}. However, current CPR rainfall data (version 5) shall be only used for detection of rainfall and using them in the context of either Bayesian or deep learning retrieval algorithms might lead to over estimation. (ii) The coincidences of active snowfall retrievals from DPR shall be used with caution in any passive microwave retrieval algorithm as it might lead to significant overestimation of snowfall. (iii) A thorough validation of satellite snowfall retrievals is essential. The ground-based observations from the National Centers for Environmental Prediction (NCEP) Automated Data Processing (ADP) and the International Comprehensive Ocean–Atmosphere Data Set (ICOADS) can be used for such a validation as suggested in \citep{Sims2015}.
A software tool, which reproduces the main results of the paper, is made publicly available at: (\url{https://github.com/aebtehaj/DIEGO})

\section*{Acknowledgment}
The research is primarily supported by grants from NASA's Remote Sensing Theory program (RST, 80NSSC20K1717) through Dr. Lucia Tsaoussi and NASA's Interdisciplinary Research in Earth Science (IDS) program (IDS, 80NSSC20K1294) through Dr. Will McCarty and Dr. Aaron Piña. Authors would like to thank Dr. F. Joseph Turk for providing CloudSat-GPM coincidence dataset, and Dr. Mircea Grecu for providing feedback on the algorithm and the results of this research paper.

\appendix
\section{Debiasing DIEGO-DPR snowfall retrievals} \label{sec:appedix}

To be able to compare DIEGO-DPR snowfall retrievals with active DPR retrievals, the snowfall estimations of DIEGO-DPR are mapped onto the 2A-DPR Ku-band footprint and the total amount of snowfall is calculated in 2021 (Fig. \ref{fig:1-Apx}). The DIEGO-DPR shows overestimation both over land and ocean compare to the active DPR retrievals because of the explained reasons in the text. In order to mitigate this bias, a latitudinal bias correction factor is calculated by dividing the zonal mean values of 2A-DPR snowfall by the DIEGO-DPR.

\setcounter{figure}{0} 
\begin{figure*}[h]
    \centering
    \includegraphics[width=\textwidth]{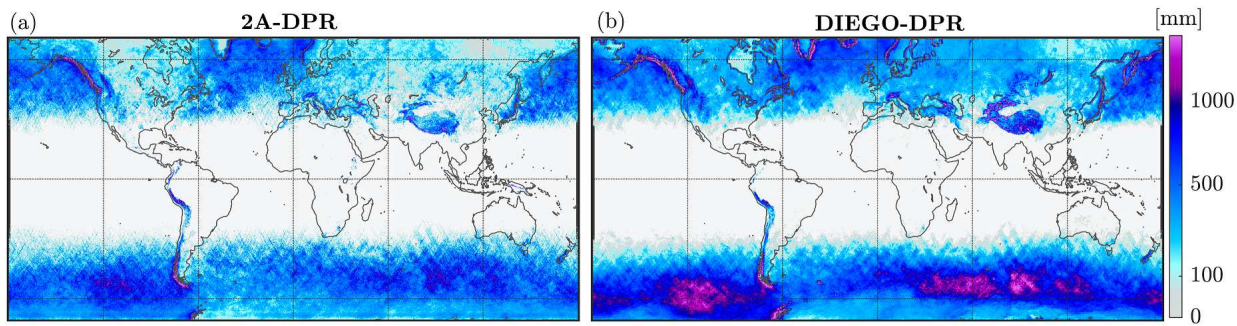}
    \caption{Annual estimates of the total snowfall in 2021 mapped onto a 0.1$^\circ$ grid box for DPR, and DIEGO-DPR products.}
    \label{fig:1-Apx}
\end{figure*}

\bibliographystyle{plainnat}
\bibliography{references.bib}

\end{document}